\pdfoutput=1

\documentclass[11pt]{article}

\usepackage{acl}

\usepackage{adjustbox}
\usepackage{amsmath}

\usepackage{times}
\usepackage{latexsym}

\usepackage[T1]{fontenc}

\usepackage[utf8]{inputenc}
\usepackage{booktabs}

\usepackage{microtype}

\usepackage{inconsolata}

\usepackage{graphicx}

\usepackage{xspace}
\usepackage{makecell}
\usepackage[dvipsnames]{xcolor}

\newcommand{\positive}[1]{\textcolor{green!60!black}{#1}}
\newcommand{\negative}[1]{\textcolor{red!70!black}{#1}}

\newcommand{\doublespace}[1]{\xspace\xspace}
\newcommand{\singlespace}[1]{\xspace}

\newcommand{\tbf}[1]{\textbf{#1}}
\newcommand{\hulm}[0]{HuLM\xspace}
\newcommand{\huft}[0]{HuFT\xspace}
\newcommand{\tft}[0]{TFT\xspace}

\newcommand{\largehlcorpus}[0]{LHLC\xspace}

\newcommand{\eos}[0]{\textit{eos}\xspace}
\newcommand{\hlunroll}[0]{HU-Llama\xspace}
\newcommand{\lhlc}[0]{Llama{\tiny{LHLC}}\xspace}
\newcommand{\llama}[0]{Llama\xspace}

\usepackage{array}
\newcolumntype{L}{>{\centering\arraybackslash}m{5cm}}

\title{Addressing the Ecological Fallacy in Larger LMs with Human Context}

\author{Nikita Soni$^1$, Dhruv Vijay Kunjadiya$^1$, Pratham Piyush Shah$^1$, Dikshya Mohanty$^1$, \\ {\bf H. Andrew Schwartz$^{1,2}$, \and Niranjan Balasubramanian$^1$} \\
$^1$Department of Computer Science, Stony Brook University \\
$^2$College of Connected Computing, Vanderbilt University, USA \\
\texttt{\{nisoni, niranjan\}@cs.stonybrook.edu}}

\begin{document}
\maketitle
\begin{abstract}
Language model training and inference ignore a fundamental linguistic fact -- there is a dependence between multiple sequences of text written by the same person. 
Prior work has shown that addressing this form of \textit{ecological fallacy} can greatly improve the performance of multiple smaller (\textasciitilde124M) GPT-based models.
In this work, we ask if addressing the ecological fallacy by modeling the author's language context with a specific LM task (called HuLM) can provide similar benefits for a larger-scale model, an 8B Llama model. 
To this end, we explore variants that process an author's language in the context of their other temporally ordered texts. We study the effect of pre-training with this author context using the HuLM objective, as well as using it during fine-tuning with author context (\textit{HuFT:Human-aware Fine-Tuning}). 
Empirical comparisons show that addressing the ecological fallacy during fine-tuning alone using QLoRA improves the performance of the larger 8B model over standard fine-tuning. Additionally, QLoRA-based continued HuLM pre-training results in a human-aware model generalizable for improved performance over eight downstream tasks with linear task classifier training alone.
These results indicate the utility and importance of modeling language in the context of its original generators, the authors.
\end{abstract}

\section{Introduction}

\begin{figure}[h!]
    \centering
    \includegraphics[width=1\linewidth]{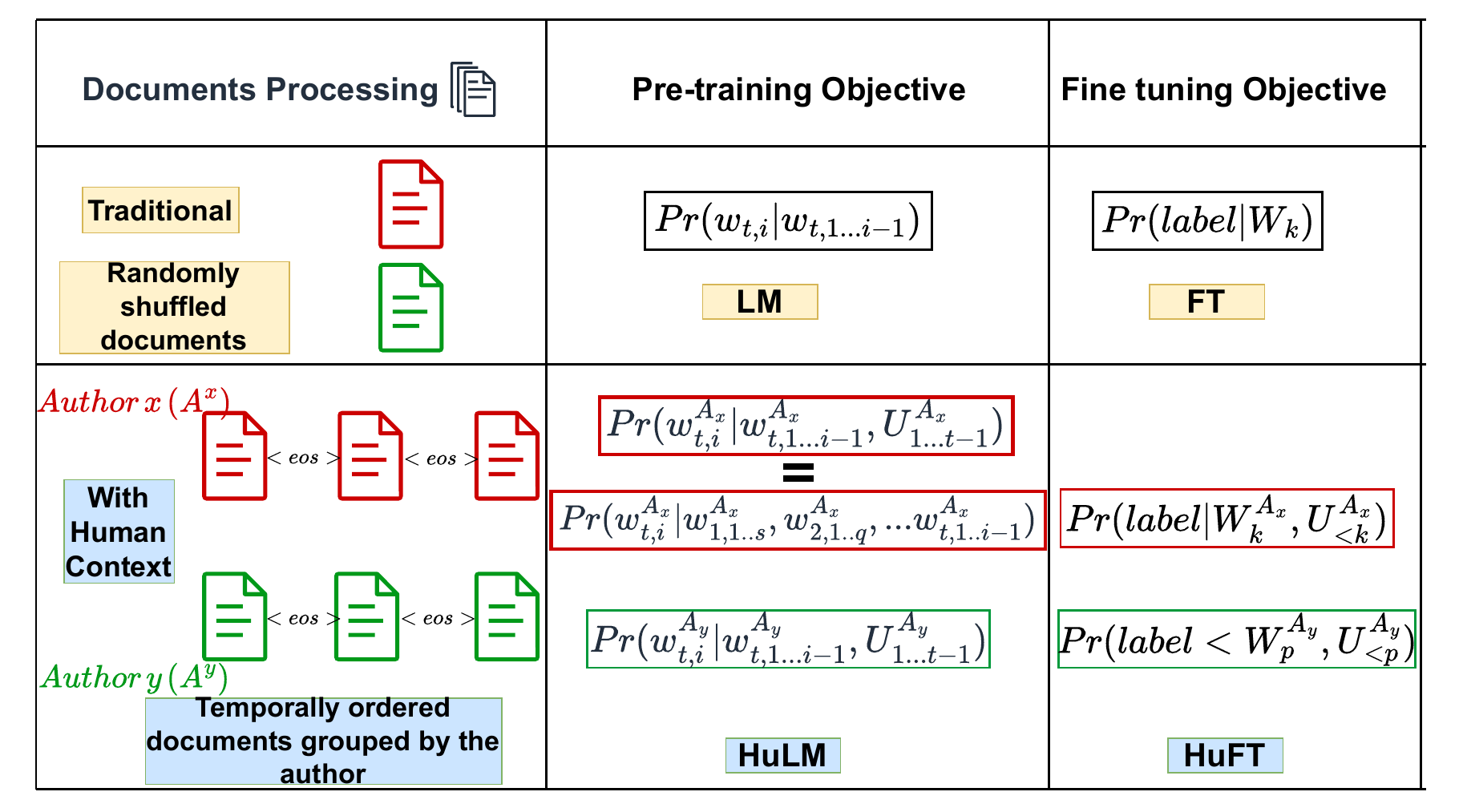}
    \caption{Addressing the \textit{ecological fallacy} in larger LMs with human context involving processing documents, pre-training, and fine-tuning within the author's context. Human Language Modeling (HuLM) and Human-aware Fine-Tuning (HuFT) consist of training objectives conditioned on author's historical language as the human context ($U$).}
    \label{fig:trad_vs_human}
\end{figure}

To date, if one asks an LLM to complete the phrase, ``Language is generated by \_\_'', they will get `humans' or `people' as the two most likely words to follow. 
Yet, the standard language modeling task itself does not model the dependence between the token sequences and the people behind language, an \textit{ecological fallacy} of assuming sequences from the same person are independent (or treated the same as those from different people). Language representations devoid of their originating human contexts lack the richness and variance that natural human contexts bring (e.g. generated language lacks variance in expressed psychological traits~\citep{giorgi2023slept, varadarajan2025consistent}), limiting model ability to address biases ~\cite{soni-etal-2024-large}.

In this work, we ask: \textit{does addressing this ecological fallacy help large language models?} In particular, we explore the impact of processing language within the human context as modeled by the author's previous texts. Prior work showed that this ecological fallacy can be remedied through a \textit{Human Language Modeling} (HuLM) task, which models the human behind the language by conditioning next-word prediction not only on the immediate discourse context but also on the broader human context via temporally ordered texts from the same author. Continued pretraining of a small scale GPT-2 variant (124M parameters) on this HuLM task improves performance both in terms of LM perplexity and downstream applications~\citep{soni-etal-2022-human}. 

However, it is not clear \emph{a priori} that the more powerful larger models need this additional human context. 
One may posit that LLMs with billions of parameters trained over trillions of tokens, already capture language from a large population of humans and thus overcome any representational or distributional shortcomings that arise from the lack of processing in the author's context. 

To address this, we investigate three different ways of incorporating human context, in terms of author's historical language, into larger LMs: (i) directly including human context to the text being processed for downstream tasks and training a task-specific linear classifier, (ii) fine-tuning model parameters using QLoRA for downstream tasks by including human context to the text, which we call \textit{HuFT: Human-aware Fine-Tuning}, and (iii) continuing pre-training model parameters using QLoRA by including human context (i.e., HuLM). We select open source \llama 3.1 8B model weights for our study, curate a new Large Human Language Corpus (LHLC) for continued \hulm pre-training, and evaluate the three ways to include human context over eight downstream tasks. We scope our study to one model family as it is a resource intensive endeavor, requiring substantial compute and time.

Our empirical results highlight several important findings: (i) Human-aware QLoRA-based Fine-Tuning (HuFT) effectively includes human context in task specific settings resulting in improved performance, (ii) Continued \hulm pre-training (QLoRA-based) results in a human-aware model that can better generalize over multiple tasks with linear classifier training alone, however, (iii) training a task-specific linear classifier by directly including the human context proves to be ineffective for non human-aware models.


In summary, we make the following contributions in this work: 1) an empirical demonstration of the value in addressing the ecological fallacy in larger language models, 2) trained a bigger HuLM model (in the range of~8B parameters) using QLoRA in multiple settings, and 3) developed a diverse and substantial HuLM data corpus consisting of texts from Reddit ~\cite{GIORGI2024116058, liu-etal-2024-aligning}, Blog Authorship Corpus ~\cite{schler2006effects}, Twitter ~\cite{giorgi2024evaluating, soni-etal-2022-human}, Gutenberg Books ~\cite{bejan2021gutenberg}, Amazon Product Reviews ~\cite{hou2024bridging}, and StackExchange ~\cite{h4stackexchange}, as well as 4) expanded tasks and dataset with author context via author's historical texts. We also present results with prompting, ablations and qualitative analysis to further support our findings.

\section{Related Work}
\label{sec:related}

A wealth of past work has shown the efficacy of looking at language within the larger context of who the author is~\citep{soni2024comparing, soni2025evaluation} or their demographics in multiple applications, such as sentiment analysis~\cite{DBLP:journals/corr/abs-2110-00135}, reducing social biases~\cite{garimella-etal-2022-demographic}, or mental health assessments~\cite{varadarajan2024archetypes, soni2025we}. In the realm of large LMs, prior work has shown benefits in considering a person's dynamic emotional states~\citep{ganesan2022wwbp, singh2025systematic} to generate empathetic dialogs~\cite{wang-etal-2022-empathetic}, or enhancing personalized responses~\cite{tan2025democratizinglargelanguagemodels} by injecting memory within model parameters using multiple LoRA modules, inspired by human memory mechanisms~\cite{zhang2025personalizedllmresponsegeneration}.

Past works~\cite{soni-etal-2022-human, soni-etal-2024-large} suggest including human context within the LM pre-training task of next word prediction to address the ecological fallacy of the LM task. \citeauthor{soni-etal-2022-human} introduce the task of \textbf{Human Language Modeling (\hulm)}, where they predict the next word given the previous words and an additional author's context in terms of the author's prior language, formulated as:

\begin{equation}
\label{eq:word_given_userstate}
Pr(\mathbf{W}_t|\mathbf{U}_{t-1}) = \prod_{i=1}^n Pr(w_{t,i} | w_{t,1:i-1}, \mathbf{U}_{1:t-1})
\end{equation}
such that the task is to predict the next word ($w_{t,i}$) of a document $W_t$, given the document's previous words ($w_{t,1:i-1}$) and a dynamic human context $U_{1:t-1}$ which models all previous words in all prior documents written by the same person. 



At the same time, larger LMs have demonstrated remarkable performance in many tasks~\cite{openai2023gpt4, hendrycks2021mmlu, jimenez2023swebench, singhal2022medqa}. However, larger LMs have not yet been evaluated for the effects of processing language within a human context (i.e., an author's historical language) when continued to pre-train or when fine-tuned for downstream tasks~\citep{matero2021melt}. In this study, we assess the impact of addressing the ecological fallacy in larger LMs by extending the concept of \hulm into larger LMs in terms of continued QLoRA pre-training and fine-tuning for downstream tasks. 

\section{Models and Methodology}


We seek to evaluate the effectiveness of processing language within the author's context, i.e., mitigating the \textit{ecological fallacy}, in larger LMs. We include this human context in three ways: i) training a task classifier using pre-trained embeddings, ii) fine-tuning model parameters for a downstream task, and iii) continue to pre-train for the human language modeling task.


We select Llama 3.1 8B ~\cite{touvron2023llamaopenefficientfoundation} as our base model. It adopts a decoder-only transformer architecture, allowing us to easily adapt it to the autoregressive \hulm task. Similar to \citet{soni-etal-2022-human}, we include the human context by collectively processing language written by the same author in the three ways we listed above. First, we discuss the difference in processing data traditionally and within the human context (i.e., within the author's context). Next, we describe the three ways that we use to include this human context and assess its impact on downstream tasks (pictorially refer Figure~\ref{fig:trad_vs_human}). Finally, we define the baseline models and details of the training setup.

\paragraph{Processing Data within the Human Context.}
Traditionally, documents are randomly shuffled and processed independently. To mitigate the ecological fallacy in language processing, we process documents generated by a common source (i.e.,  the author) together, thus inducing dependence on this hierarchical source.
In practice, we concatenate documents written by the same author, separated by the special \textit{\eos} token, ordering them temporally when the created time information is available. This approach of processing documents within the author's context remains consistent for both pre-training and fine-tuning with the human context.

For document-level labeled downstream tasks, the models process all the tokens in the concatenated sequence and use the target document's last token's hidden states (of the last layer) to predict the associated label, e.g., the author's stance or sentiment. For person-level labeled downstream tasks, the models process the author's language similarly and use the averaged token embeddings across all tokens from an author to predict/estimate the associated label, e.g., the author's occupation or age. 

\paragraph{Classifier Training with the Human Context.}
In this approach, we use pre-trained model's embeddings, precisely the last layer's hidden states, for documents processed within the author's context. These are processed in accordance to document- or person- level downstream tasks and used as inputs to a linear classifier which is trained for specific downstream tasks.


\paragraph{Human-aware Fine-Tuning: HuFT.} 
Similar to the approach of classifier training with the human context, the model's last layer's hidden states are processed based on the type of downstream task and provided as inputs to a linear classifier. However, in this method we train the classifier as well as fine-tune the model parameters for specific downstream tasks.





\paragraph{Continued HuLM Pre-training: \hlunroll.}

We continue to pre-train \llama 3.1 8B for the next word prediction task using our curated dataset: \largehlcorpus (see Section~\ref{sec:pt_data}), however, we do so over temporally-ordered documents written by a particular author, concatenated using the special \eos token, instead of randomly sampling documents and processing them independently. This results in each instance representing an author, inducing an explicit author's context by introducing dependence on the language written by the same author. This follows a flattened version of the data processing and \hulm pre-training task from ~\citet{soni-etal-2022-human}, resulting in HU-Llama (Human-aware Llama).

\paragraph{Baselines: Traditional Fine-tuning (TFT), \llama, \lhlc.} 
We use traditional fine-tuning (TFT) as the baseline to compare against HuFT. Here, we adopt the standard fine-tuning approach where the model is given an individual document and asked to predict the label. For document-level tasks, this translates to not using the author's context. For person-level tasks, we adopt a prior common approach~\cite{soni-etal-2022-human} of fine-tuning the model by independently predicting a person-level attribute for each document written by the author and taking an average or the mode of the predictions across all documents to arrive at a person-level prediction.

We compare the bigger \hulm model, \hlunroll, with their counterpart non-\hulm model, \llama, across downstream tasks using classifier training and human-aware fine-tuning (HuFT). While \hulm models are naturally designed to adopt \huft, smaller LMs are not known to use this approach, usually due to the limitation of context lengths. However, larger LMs that can process larger contexts have not been evaluated for \huft. So in addition to the \hulm model (\hlunroll), we also evaluate \llama (non-\hulm model) using \huft for downstream tasks, leveraging their capacity to process long contexts.

\begin{table}[h!]
    \centering
    \setlength{\tabcolsep}{5pt}
    \small
    \begin{adjustbox}{width=0.5\textwidth}
    \begin{tabular}{lccccccc}
        \toprule
        \textbf{Dataset} & \textbf{Epochs} & \textbf{Users} & \textbf{Docs} & \textbf{Tokens} & \textbf{UTF-8 bytes} \\
         & & & \textit{(millions)} & \textit{(millions)} & \textit{(GB)} \\
        \midrule
        Amazon        & 1   & 30,902   & 2.276 & 208.37 & 0.86 \\
        Blogs         & 3   & 19,525   & 0.322 & 91.99  & 0.36 \\
        Books         & 1   & 3,425    & 0.005 & 262.21 & 1.09 \\
        Twitter          & 3   & 20,135   & 2.414 & 66.00  & 0.24 \\
        Reddit        & 3  & 28,229   & 1.482 & 177.15 & 0.73 \\
        StackExchange & 1  & 15,440   & 0.564 & 83.69  & 0.35 \\
        \midrule
        \textbf{Total} &  & \textbf{117,656}  & \textbf{7.063} & \textbf{889.41} & \textbf{3.64} \\
        \bottomrule
    \end{tabular}
    \end{adjustbox}
    \caption{Subset of Large Human Language Corpus (LHLC) used as our pre-training data. Token counts are based on LLaMA 3.1 tokenizer.}
    \label{tab:pt_data_stats}
\end{table}

Further, to tease apart the impact of our pre-training data corpus---LHLC---, we continue to pre-train \llama 3.1 8B for the standard next word prediction task using our LHLC dataset and call it \lhlc. We compare our methodologies of including human context with \lhlc as well.




\paragraph{Training Setup: QLoRA Training, and Experimental Settings.}

For both, continued pre-training and fine-tuning the models, we use low-rank adapters and 4-bit quantized weights (QLoRA)~\cite{dettmers2023qloraefficientfinetuningquantized} accommodating for compute availability. Additionally, we use mixed-precision training, performing operations in half-precision format to speed up computation, as well as use Accelerate~\cite{accelerate} for a distributed environment in some time-consuming experiments. We use the PEFT library~\cite{han2024parameterefficient} integrated in the Hugging Face (HF) library~\cite{wolf-etal-2020-transformers} to train QLoRA weights associated with Q, K, V, O of the self-attention mechanism. 

At the 8B scale, this setup enables us to continue \hulm pre-training with a batch size of 3 per GPU on our hardware (NVIDIA H100 80GB GPUs) using 8192 tokens per author-instance. Similarly, using ditto document-tokens, we use a batch size of 123 with each instance (representing each document) limited to 200 tokens for \lhlc continued pre-training, where each document is processed independently. We run initial experiments with smaller data samples using learning rates 1e-6, 3e-4, and 5e-5, and resort to the best performing, 3e-4, when training with full data. We train on full data incrementally, and Table~\ref{tab:pt_data_stats} details the number of epochs for which each data source was trained. Finally, we merge the QLoRA pre-trained \hulm- and traditional-LM- adapters into the respective base \llama models, yielding \hlunroll and \lhlc. We use a similar QLoRA setup for HuFT and TFT starting with the resulting pre-trained models. We train all models for 5 epochs or 10 epochs with a learning rate of 3e-4 and an early stopping threshold set to 6 on the evaluation loss. We cap the training tokens to 4096 per instance (i.e., per author) and optimize training using cross-entropy loss for classification tasks and mean squared error loss for regression tasks. 

We run statistical significance testing using paired t-test for regression tasks and permutation test for classification tasks.

\section{Datasets and Tasks}

\begin{table*}[h!]
    \centering
    \setlength{\tabcolsep}{5pt}
    \small
    \begin{adjustbox}{width=\textwidth}
    \begin{tabular}{lrrrrrrrrrr}
        \toprule
        \textbf{Task} & \textbf{Users} & \textbf{Docs} & \textbf{Labels} & 
        \textbf{wpd$_{med}$} & \textbf{wpd$_{max}$} &  
        \textbf{dpa$_{med}$} & \textbf{dpa$_{max}$} & 
        \textbf{wpa$_{med}$} & \textbf{wpa$_{max}$} \\
        \midrule
        Occupation\tiny{(blogs}) & 3,539 & 47,500 & 3,539 & 95 & 350 & 8 & 337 & 1166 & 4999 \\
        Age\tiny{(blogs}) & 15,942 & 210,253 & 15,942 & 100 & 350 & 9 & 353 & 1189 & 4999 \\
        Movie Reviews\tiny{(amazon)} & 16,369 & 109,213 & 109,213 & 119 & 350 & 6 & 17 & 826 & 3783 \\
        Business Reviews\tiny{(yelp)} & 35,025 & 216,166 & 216,166 & 133 & 350 & 6 & 8 & 848 & 2090 \\
        Electronics Reviews\tiny{(amazon)} & 15,026 & 130,316 & 130,316 & 67 & 350 & 9 & 12 & 719 & 2706 \\
        Book Reviews\tiny{(amazon)} & 18,475 & 156,218 & 156,218 & 122 & 350 & 8 & 15 & 1077 & 3293 \\
        Sentiment\tiny{(twitter)} & 10055 & 47280 & 10485 & 16 & 192 & 1 & 199 & 21 & 3478 \\
        Stance\tiny{(twitter)} & 2349 & 17411 & 3021 & 15 & 33 & 1 & 2095 & 19 & 40018 \\
        
        \bottomrule
    \end{tabular}
    \end{adjustbox}
    \caption{Curated two person-level (Occupation and Age) and four document-level task datasets. Statistics reported above, where wpd = words per document, dpa = documents per author, wpa = words per author, and med = median. Stance and Sentiment are two existing document-level datasets.} 
    \label{tab:downstream_stats}
\end{table*}

\subsection{Pre-training Data: Large Human Language Corpus (\largehlcorpus)}
\label{sec:pt_data}
Human language modeling requires pre-training datasets that can provide language in the author's context, i.e., where text can be attributed to its source (author) while maintaining the privacy of a person's identity. Despite the abundant availability of datasets with meta-data consisting of $\textit{anonymous user identifiers}$, to the best of our knowledge, there is no cleaned and processed dataset available to use directly. To facilitate progress in human language modeling and personalized modeling research, we build and release the first version of our pre-training dataset, \largehlcorpus---a large, multi-source corpus containing millions of documents across more than 150K authors, and a data report containing the details of our dataset construction and design principles [REDACTED]. Briefly, \largehlcorpus creation steps include: 1) removing missing data, 2) data deduplication, 3) english filtering, 4) text formatting (encoding, URLs, etc.), 5) toxicity filtering, 6) anonymization. Here, we use a subset of this data summarized in Table~\ref{tab:pt_data_stats}. We consider a mix of domains with a stronger focus on social language (e.g., blogs, Reddit, Twitter) along with more topic-focused domains (e.g., Amazon Products Reviews, StackExchange) and domains with less informal language (e.g., books) to regularize training. Details of the training split can be found in the Appendix Table~\ref{tab:pt_data_split_utf8}.

\subsection{Downstream Datasets and Tasks}

We consider eight downstream tasks that involve assessing person-level attributes grouped into two categories based on the prediction level: \textbf{document-level} and \textbf{person-level}. In the first category, given a target text sequence (document) written by a person, the model predicts a label associated with that document (e.g., the person's stance on a topic such as \textit{atheism}). In the second, given multiple text sequences (documents) written by the same person, the model predicts/estimates a label representing a broader personal attribute (e.g., occupation or age).


We select existing datasets as well as curate new datasets (similar to LHLC curation) for downstream tasks (refer Table~\ref{tab:downstream_stats}) that evaluate the impact of including the human context while processing language. These tasks
have real-world applications such as sentiment analysis, bias mitigation, enforcing safety policies (e.g., flagging underage users in sensitive domains), tailoring educational content, analyzing labor market trends, and authorship attribution. At the same time, we acknowledge the potential dual-use risks of such modeling and person associated tasks---such as profiling, manipulation, stereotyping, or targeted advertising---and emphasize the need for data consent and transparency. We discuss these considerations in greater detail in Section~\ref{sec:ethical}. Training splits can be found in Appendix Table~\ref{tab:down_split_stats}



\paragraph{Person-Level.} Similar to \largehlcorpus corpus, we curate two person-level downstream task datasets and training splits from existing blogs~\cite{schler2006effects} corpus. One requires classifying a person's \textbf{occupation}, and the other requires estimating a person's \textbf{age}, given the blogs written by them for both tasks. We limit the occupation classification data to consist of the top 5 occupations from the blogs corpus (student, technology, arts, communication and media, and education). 


\begin{table*}[h!]
    \centering
    \setlength{\tabcolsep}{5pt} 
    \renewcommand{\arraystretch}{1.1} 
    \begin{small}
    \begin{adjustbox}{width=\textwidth} 
    \begin{tabular}{l|l|cccccc|c|c}

        \hline
        
        \tbf{} & \tbf{} 
        & \multicolumn{6}{c|}{\tbf{Document-Level}} 
        & \multicolumn{2}{c}{\tbf{Person-Level}} \\

        \tbf{} & \tbf{} 
        & \multicolumn{6}{c|}{\tbf{Cls ($F1$)}} 
        & \multicolumn{1}{c|}{\tbf{Cls ($F1$)}} 
        & \multicolumn{1}{c}{\tbf{Reg ($r$)}}\\
        
        \tbf{Model} & \textbf{\makecell{Include \\ HC}}
        & \tbf{Movie} & \tbf{Business} 
        & \tbf{Book} & \tbf{Elec}
        & \tbf{Stn} & \tbf{Sent} 
        & \tbf{Occ} 
        & \tbf{Age} \\
        
        \hline

         \llama & No
    & 57.99 & 65.99 & 59.90 & 59.06
    & 64.29 & 62.01
    & 47.48
    & 0.826 \\

    \hline

    \llama & Yes
    & 55.06 & 63.37 & 57.80 & 57.61
    & 62.96 & 61.40 
    & 53.59
    & 0.853 \\

    \textbf{\hlunroll} & \textbf{Yes}
    & \textbf{\doublespace{}\underline{58.48}$^\dagger$} & 
    \textbf{\singlespace{}\doublespace{}\underline{66.80}$^{*\dagger}$} & 
    \textbf{\singlespace{}\doublespace{}\underline{61.88}$^{*\dagger}$} & 
    \textbf{\singlespace{}\doublespace{}\underline{60.93}$^{*\dagger}$} & 
    \textbf{\doublespace{}\underline{67.08}$^{\dagger}$} & \textbf{\doublespace{}\underline{62.93}$^{\dagger}$} 
    & \textbf{\doublespace{}\underline{54.73}$^*$}
    & \textbf{\singlespace{}\doublespace{}\underline{0.858}$^{*\dagger}$}  \\

        \hline
    \end{tabular}
    \end{adjustbox}
    \end{small}
    \caption{Evaluating directly including human context (HC) using pre-trained embeddings and training task-specific linear classifiers. Results are reported in weighted F1 for classification (cls) tasks and in pearson r for regression (reg) task. Bold indicates best in column, * and $\dagger$ indicate statistical significance $p < .05$ w.r.t not including human context, and w.r.t including human context using \llama embeddings respectively . Underline indicates where continued QLoRA \hulm pre-training additionally helps.}
    
    \label{tab:cls_ft_results}
\end{table*}

\begin{table*}[h!]
    \centering
    \setlength{\tabcolsep}{5pt} 
    \renewcommand{\arraystretch}{1.1} 
    \begin{small}
    \begin{adjustbox}{width=\textwidth} 
    \begin{tabular}{l|l|cccccc|c|c}

        \hline
        
        \tbf{} & \tbf{} 
        & \multicolumn{6}{c|}{\tbf{Document-Level Labels}} 
        & \multicolumn{2}{c}{\tbf{Person-Level Labels}} \\


        \tbf{} & \tbf{} 
        & \multicolumn{6}{c|}{\tbf{Cls ($F1$)}} 
        & \multicolumn{1}{c|}{\tbf{Cls ($F1$)}} 
        & \multicolumn{1}{c}{\tbf{Reg ($r$)}}\\
        
        \tbf{Model} & \textbf{FT style}
        & \tbf{Movie} & \tbf{Business} 
        & \tbf{Book} & \tbf{Elec}
        & \tbf{Stn} & \tbf{Sent} 
        & \tbf{Occ} 
        & \tbf{Age} \\
        
        \hline

        \llama  & TFT 
        & 66.40 & 70.93 & 65.32 & 65.26
        & \textbf{71.37} & \textbf{76.44}
        & 52.45
        & 0.882 \\

        \hline

        \llama{} & \bf \huft 
        & \textbf{\doublespace{}67.52$^*$} & 
        \textbf{\doublespace{}\singlespace{}74.36$^{*\dagger}$} &
        69.22 & \textbf{\doublespace{}\singlespace{}68.84$^{*\dagger}$} 
        & 69.09 & 76.07 
        & 52.28
        & \textbf{\doublespace{}0.916$^{*}$}  \\
        
        \hlunroll & \bf \huft 
        & 67.05 & 73.36  & \textbf{\doublespace{}\underline{69.43}$^{*}$} & 68.22
        & 70.77 & 75.25 
        & \textbf{\doublespace{}\singlespace{}\underline{57.50}$^{*\dagger}$}
        & \textbf{\doublespace{}0.916$^*$} \\

        \hline
    \end{tabular}
    \end{adjustbox}
    \end{small}
    \caption{Evaluating QLoRA based Human-aware Fine-Tuning (HuFT). Results reported in weighted F1 for classification (cls) and in pearson r for regression (reg) tasks. Bold indicates best in column, * and $\dagger$ indicate statistical significance $p < .05$ w.r.t traditional fine-tuning (TFT), and w.r.t second best HuFT results respectively. Underline indicates where continued QLoRA \hulm pre-training additionally helps.}
    \label{tab:qlora_ft_results}
\end{table*}

\paragraph{Document-Level.} We curate four document-level downstream task datasets from existing sources---Amazon movie Reviews~\cite{mcauley2013amateurs}, Yelp Business Reviews~\cite{DBLP:journals/corr/Asghar16}, Amazon Electronics Reviews~\cite{hou2024bridging}, and Amazon Books Reviews~\cite{hou2024bridging}. All four tasks involve predicting a rating (ranging 1 to 5) that a person would give corresponding to a review written by them. We select different domains and products to diversify the task and assess performance impact across domains and topics. Further, we select two publicly available datasets consisting of authors' context~\cite{soni-etal-2022-human} for the \textbf{stance detection} and \textbf{sentiment analysis} tasks from SemEval~\cite{nakov-etal-2013-semeval, mohammad2016semeval} using ditto train, validation, and test splits. There are five sub-datasets corresponding to five topics (atheism, abortion, climate, feminism, and Hillary Clinton) for the stance detection task, which requires predicting the stance (for/against/neutral) of a person for a particular topic given a tweet written by them. Similarly, for sentiment analysis, the task is to predict the sentiment (positive/negative/neutral) of a person given a tweet written by them. For both tasks, the author's context,  where available, consists of historically written (unlabeled) tweets by the author. 





\section{Results and Discussion}

In this section, first we describe the main results of studying the impact of incorporating human context in larger LMs in three settings: (i) classifier-only training, (ii) QLoRA-based Human-aware Fine-Tuning (HuFT), and (iii) QLoRA-based continued \hulm pre-training, presented in Tables~\ref{tab:cls_ft_results} and ~\ref{tab:qlora_ft_results}. Next, we summarize additional results supporting our main investigative study with ablations, interpretive experimental and qualitative analyses. Finally, in section~\ref{sec:results_discussion} we discuss the inferences and implications for the observations stated. 



\subsection{Including human context with task-specific classifier-only training}
We find that directly including human context using \llama embeddings, for task-specific classifier-only training, improves performance on person-level tasks (estimating age and classifying occupation). However, it does not benefit any of the document-level tasks (see Table~\ref{tab:cls_ft_results}, rows ``\llama with human context'' and ``\llama without human context''). At the same time, directly including human context using \hlunroll embeddings---which is \llama continued to pre-train with human context for \hulm task using QLoRA---for task-specific classifier training yields substantial gains across all the document- and person-level tasks. Five out of eight tasks, including both person-level tasks, show statistically significant results when including human context as compared to not including human context.

\subsection{Including human context in downstream task fine-tuning: HuFT}

We find that human-aware task-specific fine-tuning \llama using QLoRA (see \huft in Table ~\ref{tab:qlora_ft_results}) shows substantial performance gains over fine-tuning without human context (\tft) in six out of eight downstream tasks with statistical significance ($p < .05$). Two document-level tasks---stance and sentiment---does not benefit in HuFT setting, however, have no statistically significant difference from TFT setting. We also note here that these two tasks have only a few instances with historical author context. 

\begin{table*}[h!]
    \centering
    \setlength{\tabcolsep}{5pt} 
    \renewcommand{\arraystretch}{1.1} 
    \begin{small}
    \begin{adjustbox}{width=\textwidth} 
    \begin{tabular}{l|l|cccccc|c|c}

        \hline
        
        \tbf{} & \tbf{} 
        & \multicolumn{6}{c|}{\tbf{Document-Level Labels}} 
        & \multicolumn{2}{c}{\tbf{Person-Level Labels}} \\


        \tbf{} & \tbf{} 
        & \multicolumn{6}{c|}{\tbf{Cls ($F1$)}} 
        & \multicolumn{1}{c|}{\tbf{Cls ($F1$)}} 
        & \multicolumn{1}{c}{\tbf{Reg ($r$)}}\\
        
        \tbf{Model} & \textbf{FT style} 
        & \tbf{Movie} & \tbf{Business} 
        & \tbf{Book} & \tbf{Elec}
        & \tbf{Stn} & \tbf{Sent} 
        & \tbf{Occ} 
        & \tbf{Age} \\
        
        \hline

         \llama & TFT
    & 57.99 & 65.99 & 59.90 & 59.06
    & 64.29 & 62.01
    & 47.48
    & 0.826 \\

    \lhlc & TFT
    & 58.31 & \doublespace{}65.64$^{o}$ & \doublespace{}61.18$^{o}$ & \doublespace{}60.22$^{o}$
    & \textbf{\doublespace{}69.25$^o$} & \textbf{63.17} 
    & \doublespace{}50.61$^o$
    & \doublespace{}0.835$^o$ \\


    \hline

    \llama & \textbf{HuFT}
    & 55.06 & 63.37 & 57.80 & 57.61
    & 62.96 & 61.40 
    & 53.59
    & 0.853 \\

    \lhlc & \textbf{HuFT}
    & 55.43 & 63.57 & \singlespace{}59.36 $\hat{}$ & 57.54
    & \singlespace{}66.37 $\hat{}$ & 62.38 
    & 54.01
    & \singlespace{}0.855 $\hat{}$ \\

    \hlunroll & \textbf{HuFT}
    & \textbf{\doublespace{}58.48$^\dagger$} & 
    \textbf{\doublespace{}\singlespace{}66.80$^{*\dagger}$} & 
    \textbf{\doublespace{}\singlespace{}61.88$^{*\dagger}$} & 
    \textbf{\doublespace{}\singlespace{}60.93$^{*\dagger}$} & 
    67.08 & 62.93 
    & \textbf{\doublespace{}54.73$^*$}
    & \textbf{\doublespace{}0.858$^*$} \\

        \hline
    \end{tabular}
    \end{adjustbox}
    \end{small}
    \caption{Evaluating the effect of our LHLC data corpus on directly including human context using task classifier-only training. Results reported in weighted F1 and pearson r. Bold indicates best in column. We indicate statistically significant ($p < .05$) difference in results between: \lhlc HuFT and \hlunroll HuFT with $^\dagger$, \lhlc TFT and \hlunroll HuFT using $^*$, \lhlc TFT and \llama TFT with $^0$, and \lhlc HuFT and \llama HUFT using $\hat{}$.}
    
    \label{tab:lhlc_cls_results}
\end{table*}

\begin{table*}[h!]
    \centering
    \setlength{\tabcolsep}{5pt} 
    \renewcommand{\arraystretch}{1.1} 
    \begin{small}
    \begin{adjustbox}{width=\textwidth} 
    \begin{tabular}{l|l|cccccc|c|c}

        \hline
        
        \tbf{} & \tbf{} 
        & \multicolumn{6}{c|}{\tbf{Document-Level Labels}} 
        & \multicolumn{2}{c}{\tbf{Person-Level Labels}} \\


        \tbf{} & \tbf{} 
        & \multicolumn{6}{c|}{\tbf{Cls ($F1$)}} 
        & \multicolumn{1}{c|}{\tbf{Cls ($F1$)}} 
        & \multicolumn{1}{c}{\tbf{Reg ($r$)}}\\
        
        \tbf{Model} & \textbf{\makecell{Include \\ HC}}
        & \tbf{Movie} & \tbf{Business} 
        & \tbf{Book} & \tbf{Elec}
        & \tbf{Stn} & \tbf{Sent} 
        & \tbf{Occ} 
        & \tbf{Age} \\
        
        \hline

         \llama-Instruct & No
    & \textbf{\doublespace{}61.14$^*$} & 59.17 & \textbf{\doublespace{}60.16$^*$} & 59.72 & 68.44 & \textbf{\doublespace{}65.75$^*$}
    & - & -
     \\

    \hline

    \llama-Instruct & Yes
    & 59.70 & \textbf{\doublespace{}60.34$^*$} & 58.63 & \textbf{\doublespace{}61.16$^*$} & \textbf{\doublespace{}70.35$^*$} & 64.81
    & 37.09 & 0.145
     \\



        \hline
    \end{tabular}
    \end{adjustbox}
    \end{small}
    \caption{Evaluating directly including human context (HC) in prompting using \llama-Instruct 3.1 8B model. Results reported in weighted F1 for classification (cls) and in pearson r for regression (reg) tasks. Bold indicates best in column, * indicates statistical significance $p < .05$ between not including human context and including human context for task-specific prompting.
    }
    \label{tab:zero_shot_prompt_results}
\end{table*}

\subsection{Including human context in continued pre-training: HuLM}
While continued QLoRA \hulm pre-training in task-specific classifier-only training setting shows benefits (see \hlunroll in Table ~\ref{tab:cls_ft_results}), we also see that these gains are surpassed by \llama with human-aware QLoRA task fine-tuning in three tasks (Movie, Business, and Electronics Reviews) with two having statistically significant results (see Table~\ref{tab:qlora_ft_results}). Two other tasks (occupation, Book Reviews Rating) sustain best performances with \hlunroll in \huft settings and one other (estimating age) is at par with \llama \huft.


\subsection{Ablation and Interpretive Analyses}

\paragraph{Effect of LHLC dataset.}
Training on a large dataset, such as LHLC, could itself explain part of the performance gains we observe on downstream tasks with \hlunroll in classifier-only training setting. Table~\ref{tab:lhlc_cls_results} shows continued pre-training (QLoRA-based) on LHLC for the standard LM task (\lhlc) fares worse compared to \hlunroll on all eight downstream tasks in the \huft setting, and on six tasks in the TFT setting. Conversely, \lhlc shows better or at par performance as compared to \llama in both \tft and \huft settings showing the benefits of continued LM pre-training on LHLC.

We perform a similar analysis to evaluate the effect of LHLC on QLoRA-based HuFT setting and report results in the Appendix Table~\ref{tab:lhlc_qlora_results} with similar findings except minor mixed results for \lhlc verus \llama, albeit, following performance trends as seen in Table~\ref{tab:qlora_ft_results}.


\begin{table*}[h!]
    \centering
    \setlength{\tabcolsep}{5pt}
    \begin{small}
    \begin{adjustbox}{width=\textwidth} 
    \begin{tabular}{l|l|ccc|ccc|ccc|ccc}

        \hline
        
        \tbf{} & \tbf{} 
        & \multicolumn{6}{c|}{\tbf{Document-Level}} 
        & \multicolumn{6}{c}{\tbf{Person-Level}} \\

        \tbf{} & \tbf{} 
        & \multicolumn{6}{c|}{\tbf{Cls ($F1$)}} 
        & \multicolumn{3}{c|}{\tbf{Cls ($F1$)}} 
        & \multicolumn{3}{c}{\tbf{Reg ($r$)}}\\
        
        
        \bf{Model} & FT style & \multicolumn{3}{c|}{\bf{Business}\tiny{(F1)}} & \multicolumn{3}{c|}{\bf{Sentiment}\tiny{(F1)}} 
        & \multicolumn{3}{c|}{\bf{Occ}\tiny{(F1)}} 
        & \multicolumn{3}{c}{\bf{Age}\tiny{(F1)}}\\
        
         \tbf{} & &  
         \bf{RS\tiny{42}} & \bf{RS\tiny{3}} & \bf{RS\tiny{1234}} 
         & \bf{RS\tiny{42}} & \bf{RS\tiny{3}} & \bf{RS\tiny{1234}} 
         & \bf{RS\tiny{42}} & \bf{RS\tiny{3}} & \bf{RS\tiny{1234}} 
         & \bf{RS\tiny{42}} & \bf{RS\tiny{3}} & \bf{RS\tiny{1234}} \\

         \hline

         \llama{} & TFT & 70.93 & 71.70 & 71.24 
         & \textbf{76.44} & 76.35 & \textbf{\doublespace{}79.05}$^{*}$ 
         & 52.45 & 50.71 & 50.21 
         & 0.882 & 0.881 & 0.885 \\

         \hline
         
         \llama & HuFT & \textbf{\doublespace{}\singlespace{}74.36}$^{*\dagger}$ & \textbf{\doublespace{}\singlespace{}74.12}$^{*\dagger}$ & 73.71  
         & 76.07 & 76.86 & 76.44 
         & 52.28 & 55.65 & 55.34 
         & \textbf{\doublespace{}0.916}$^{*}$ & \textbf{\doublespace{}0.913}$^{*}$ & \textbf{\doublespace{}0.914}$^{*}$ \\
         \hlunroll & HuFT & 73.36 & 73.48 & \textbf{\doublespace{}\singlespace{}74.19}$^{*\dagger}$ 
         & 75.25 & \textbf{\doublespace{}78.05}$^{*}$ & 75.32 
         & \textbf{\doublespace{}\singlespace{}57.50}$^{*\dagger}$ & \textbf{\doublespace{}56.76}$^{*}$ & \textbf{\doublespace{}55.82}$^{*}$ 
         & \textbf{\doublespace{}0.916}$^{*}$ & \textbf{\doublespace{}0.913}$^{*}$ & \textbf{\doublespace{}0.914}$^{*}$ \\
         
         \hline
    
    \end{tabular}
    \end{adjustbox}
    \end{small}
    \caption{QLoRA-based \huft and \tft experiments with different random seeds (RS) for selected downstream tasks. Results in weighted F1 for classification (cls) tasks and in pearson r for regression (reg) task. Bold indicates best in column, * and $\dagger$ indicate statistical significance $p < .05$ w.r.t \tft and w.r.t second best HuFT results respectively.}
    \label{tab:random_seed_qlora_results}
\end{table*}

\paragraph{Including human context in prompting.}
We further evaluate directly including human context in prompting using \llama-Instruct 3.1 8B model (prompts and details in Appendix~\ref{app:prompt}). We find mixed results in directly including author's historical language with zero-shot prompting (see Table~\ref{tab:zero_shot_prompt_results}\footnote{Since it is not natural to prompt for person-level tasks with individual texts (i.e., no human context), we skip those in the results table.}. We also experiment with limited historical context and historical context along with labels over two selected downstream tasks but notice only marginal differences (see Appendix Table~\ref{tab:zero_shot_prompt_ablations}). Overall, \llama-Instruct 3.1 8B model is not able to effectively use the human context with prompting.


\paragraph{Experiments with randomness.}
We experiment with two other random seeds ($3$ and $1234$) to observe the effect of randomness on selected downstream tasks. We find similar performance results as our original random seed ($42$) on selected tasks with an exception where \huft performs better than \tft for sentiment analysis with one random seed (see Table~\ref{tab:random_seed_qlora_results}). These results are consistent with the main findings of \huft improving downstream task performance regardless of the randomness.

\begin{table*}[h!]
    \centering
    \setlength{\tabcolsep}{5pt}
    \small
    \begin{adjustbox}{width=\textwidth}
    \begin{tabular}{p{2cm}|L|L|L}
    \hline
    \textbf{Task} &
    \textbf{Target Text} &
    \textbf{Selected Human Context} &
    \textbf{Why Human Context Helps/Hurts} \\
    \hline

     \makecell*[l]{
     Stance (Atheism) \\
     \positive{True: FOR}\\ \positive{HuFT: FOR} \\ \negative{TFT: AGAINST}} 

    &
    I \negative{\#DenounceHarper} for 
    refusing  to include 
    \negative{family planning} 
    in foreign aid even though spending \$1  
    could save \$6
    \negative{\#wherestheFP} 
    
    &

     [...] bringing in
    \positive{religious groups [...]
    to access our kids in secrecy}. [\dots]
    \positive{anti-choice [..] denying accurate SexED}
    [..] harm they cause. [\dots] 
    &

    The user’s history repeatedly links policy to
    \positive{religious influence} and \positive{Sex ED},
    clarifying the stance. 
    
    \\
    \hline

    \makecell*[l]{%
    Occupation \\
    \positive{True: Tech.} \\
    \positive{HuFT: Tech.} \\
    \negative{TFT: Arts}%
    } 
    
    &
    
    ``\dots\ homeward bound :: need my 
    hermit shell \dots'' \textbf{[TFT $\rightarrow$ Technology]} [...] 
    ``[\dots] 
    but \negative{the greatest  [\dots] CPU [\dots] is the human
    brain [\dots] how you 
    loop them} [\dots]''
    \textbf{[TFT $\rightarrow$ Arts]}
    &

    i am mack, i will some day become great [\dots]
    unused \positive{CPU power} [\dots]
    banning \positive{evolution from science classes} [\dots]
    reactions to \positive{NASA discoveries} and scientific progress [\dots] 
    &

    TFT can be right sometimes but majorly predicts ``arts''.
    Repeated references to \positive{computing metaphors}, \positive{science education}, and
    \positive{technology--policy reasoning} help HuFT. 
    \\
    \hline

      \makecell*[l]{%
     Business review \\
    \positive{True: 4.0} \\
    \negative{HuFT: 1.0} \\
    \positive{TFT: 4.0}%
    } 
    &

    I’ve never had Chinese food before, 
    and after the first bite, 
    \positive{``wow this tastes like it was 
    just cooked for me.''} 
    This happened every time 
    [...] 
    &

    I would \negative{never eat here}. (\negative{0.0}) [\dots] 
    \negative{no thanks}. (\negative{1.0}) [\dots] 
    \negative{a lot really pissed me off}. (\negative{2.0}) 
    & 
    History is dominated by \negative{long, critical complaints} , 
    leading the model to \negative{discount concise praise} and predict
    too low. 
    \\
    \hline

    \end{tabular}
    \end{adjustbox}
    \caption{Selected examples for qualitative analysis that demonstrate how including author's historical language as human context can improve model predictions, and in some cases, mislead them, along with plausible reasons.}
    \label{tab:qa_main_examples}
\end{table*}

\paragraph{Qualitative Analysis}
We follow a structured approach by looking at error in predictions (detailed process listed in Appendix~\ref{app:qa_process}) to manually review examples where human context as the author's historical language helps or hurts a model's predictions. We find examples where models in \tft setting make the wrong prediction because the target text does not directly reveal the stance or occupation of a person whereas models in \huft setting benefit from the historical human context to predict correctly. Conversely, we also find cases where the historical language can be misleading causing the models to mispredict in the \huft setting. We show a few examples in Table~\ref{tab:qa_main_examples} and provide task-wise qualitative analysis in the Appendix~\ref{app:qa_process}.

\subsection{Discussion}
\label{sec:results_discussion}
Human-aware QLoRA task fine-tuning proved effective for improving performance by including human context for task-specialized models where model parameter tuning is feasible.
Conversely, including human context in continued pretraining, even within the QLoRA environment, yielded a human-aware model that generalized better across multiple tasks with linear classifier training alone. At the same time, directly including human context in the classifier-only training setting was ineffective for non-human-aware models, consistent with the relatively poorer results we observe when prompting with human context. 
We hypothesize that the reason could be the inability of \llama 3.1 8B model to effectively use a large amount of human context to improve its performance on downstream tasks. This is further supported by the better results of \llama in TFT setting for the stance and sentiment analysis tasks, which are the only two tasks where sufficient author historical context is not available. Additionally, as our qualitative analysis showed, authors' historical context can be ambiguous in some instances. Such results suggest a future research study on retrieving relevant historical language to provide an effective human context.

These findings should be taken in light of several limitations: the acutely low number of model parameters (\textasciitilde0.17\%) that could be trained within the QLoRA setup, as well as the smaller size of our LHLC dataset relative to that of the original \llama pre-training data. In addition, our study comprehensively experiments with a specific family and size of models, in part because these experiments are extremely resource intensive, requiring substantial compute and time. This calls upon future studies to experiment with full or higher percentages of model parameter tuning, and to assess different model families and sizes. Further, we highlight the importance of smaller sized (\textasciitilde1B, \textasciitilde8B) models becoming human-aware, as many use cases involve sensitive data and smaller models can be hosted locally, thereby preserving user consent and privacy.






\section{Conclusion}
Scaling has delivered impressive advances for language models. 
However, these models ignore the larger dependence between sequences of text that come from the same person. 
This work studied the impact of remedying this issue in large-scale language models (with 8B parameters) by modeling the author's prior language contexts.
A simple change to the target task fine-tuning, where we incorporate the author's prior language, led to significant improvements over standard ways of fine-tuning for task-specialized models. Pretraining with author context based language modeling (\hulm) on our curated Large Human Language Corpus (LHLC) yields a human-aware model (~8B parameters) that can provide generalized benefits over multiple tasks with simply training a linear task classifier. These results together demonstrate the utility of modeling the primary generators of language, humans, in large language models.

\section*{Limitations}
\label{sec:limitations}
The purpose of our study is to consider the effects of processing language within the author's context in larger LLMs within the scope of continued pre-training and fine-tuning. We resort to quantized low rank adaptation of some model parameters as we are limited by the compute availability. This may result in reduced efficacy of the continued pre-training of the \hulm task within larger LMs. Thus, we note that assessing the full impact of \hulm pre-training in larger LMs remains an open question.
Additionally, we note that the author's context may be dependent on the quality of the text documents used from their previously written language. This is a yet another research question remaining to be explored and beyond the scope of our study.
Furthermore, our study's scope does not include prompt engineering or comprehensive assessment of the efficacy of different prompting strategies in various conditions with the author's context. We include basic prompting experiments with selected experiments on limited historical context and labeled historical context for the sake of completeness of comparison only. 

\section*{Ethical Considerations}
\label{sec:ethical}
The multi-level human-document-word architecture of \hulm enables large language models to incorporate dependencies across an individual user's prior language, rather than treating each text sample in isolation. This shift toward modeling the human generators of language unlocks new potential for improving fairness, personalization, and contextual understanding. However, the same capability that allows for richer user-level context also raises important ethical concerns---particularly regarding the risks of misuse, such as behavioral profiling or manipulation based on language history.

To mitigate these risks, we systematically review each dataset incorporated into the corpus, identifying and removing user identifiers. This process was followed by thorough manual checks to ensure that no personally identifiable information remained. These safeguards were essential for protecting user privacy and reducing the likelihood of unintended exposure of sensitive information from social media content.

Additionally, our models/architecture doesn't explicitly rely on or encode user attributes during pre-training. By focusing solely on patterns in language use—rather than incorporating static user-level features—we aim to preserve privacy while still capturing the richness of human communication. This approach aligns with our broader objective of building ethically responsible, human-centered language models.




\bibliography{custom}

\appendix
\label{sec:appendix}
\section{Appendix}

\subsection{Dataset Training Splits}

\begin{table*}[t]
\centering
\small
\begin{tabular}{lrrrrrrrrr}
\toprule
\textbf{Dataset} & 
\multicolumn{3}{c}{\textbf{Train}} &
\multicolumn{3}{c}{\textbf{Dev}} &
\multicolumn{3}{c}{\textbf{Test}} \\
\cmidrule(r){2-4} \cmidrule(r){5-7} \cmidrule(r){8-10}
 & \textbf{Users} & \textbf{Docs (M)} & \textbf{Tokens (M)}
 & \textbf{Users} & \textbf{Docs (M)} & \textbf{Tokens (M)}
 & \textbf{Users} & \textbf{Docs (M)} & \textbf{Tokens (M)} \\
\midrule
Amazon        
& 30,000 & 2.217 & 201.85
& 402 & 0.026 & 2.86
& 500 & 0.034 & 3.66 \\

Blogs         
& 17,760 & 0.293 & 83.13
& 780 & 0.013 & 3.91
& 985 & 0.016 & 4.95  \\

Books         
& 2,984 & 0.005 & 255.38
& 194 & 0.0002 & 3.01
& 247 & 0.0003 & 3.82 \\

CTLB          
& 18,393 & 2.194 & 59.72
& 766 & 0.097 & 2.76
& 976 & 0.123 & 3.52 \\

Reddit        
& 25,550 & 1.338 & 158.63
& 1,193 & 0.064 & 8.25
& 1,486 & 0.080 & 10.26 \\

StackExchange 
& 14,980 & 0.549 & 81.31
& 204 & 0.006 & 1.04
& 256 & 0.008 & 1.35 \\

\midrule
\textbf{Total}
& \textbf{109,667} & \textbf{6.622} & \textbf{840.02}
& \textbf{3,539} & \textbf{0.206} & \textbf{21.84}
& \textbf{4,450} & \textbf{0.260} & \textbf{27.55}\\
\bottomrule
\end{tabular}
\caption{Train/Dev/Test split of the Large Human Language Corpus (LHLC) used for pre-training. Token counts are based on the LLaMA 3.1 tokenizer. Number of documnets and tokens are reported in millions.}
\label{tab:pt_data_split_utf8}
\end{table*}

\begin{table*}
    \centering
    \small
    \begin{tabular}{lrrrrrrrrrr}
        \toprule
        \textbf{Task} & 
        \multicolumn{3}{c}{\textbf{Train}} & 
        \multicolumn{3}{c}{\textbf{Dev}} & 
        \multicolumn{3}{c}{\textbf{Test}} \\
        \cmidrule(r){2-4} \cmidrule(r){5-7} \cmidrule(r){8-10}
        \textbf{} & \textbf{Users} & \textbf{Docs} & \textbf{Labels} 
                 & \textbf{Users} & \textbf{Docs} & \textbf{Labels} 
                 & \textbf{Users} & \textbf{Docs} & \textbf{Labels} \\
        \midrule
        Occupation\tiny{(blogs}) & 2,135 & 28,199 & 2,135 & 532 & 7,770 & 532 & 872 & 11,531 & 872 \\
        
        Age\tiny{(blogs}) & 10,354 & 135,594 & 10,354 & 2,402 & 32,773 & 2,402 & 3,186 & 41,886 & 3,186 \\
        Movie Reviews\tiny{(amazon)} & 10,621 & 71,338 & 71,338 & 2,460	& 16,168 & 16,168 & 3,288 & 21,707 & 21,707 \\
        Business Reviews\tiny{(yelp)} & 22,856 & 141,153 & 141,153 & 5,233	& 32,179 & 32,179 & 6,936 & 42,834 & 42,834 \\
        Electronics Reviews\tiny{(amazon)} & 9,768 & 84,972 & 84,972 & 2255 & 19,527 & 19,527 & 3,003 & 25,817 & 25,817 \\
        Book Reviews\tiny{(amazon)} & 11,974 & 101,614 & 101,614 & 2,811	 & 23,565 & 23,565 & 3,690	 & 31,039 & 31,039 \\
        Sentiment\tiny{(twitter)} & 6,246 & 28,808 & 6,461 & 1,000 & 4,548 & 1,030 & 2,859 & 13,924 & 2,994 \\
        Stance\tiny{(twitter)} & 1,361 & 11,318 & 1,658 & 332 & 1,996 & 418 & 768 & 4,097 & 945 \\
        
        \bottomrule
    \end{tabular}
    
    \caption{Train, Dev, and Test split statistics for each dataset across tasks, including number of users, documents, and labels. Here, we use the splits from SemEval tasks for stance and sentiment~\cite{nakov-etal-2013-semeval, mohammad2016semeval}, and the author's context from \citet{lynn2019tweet, soni-etal-2022-human}. For the person-level tasks, we stratify on the number of words per user and maintain a consistent label-proportions and no overlapping authors in each split.}
    \label{tab:down_split_stats}
\end{table*}

\subsection{Additional Analysis}

\paragraph{Effect of author's historical context on downstream tasks.} 
Here, we perform ablations by running each model---\llama, \lhlc, and \hlunroll---in both settings, \huft (i.e., with author's historical context) and \tft (i.e., with no human context), under the QLoRA FT setup. We find that adding the human context shows consistent benefits in the Table~\ref{tab:no_hc_ablations_qlora} across all downstream tasks for all models except in two specific instances alone, involving stance and sentiment detection tasks.

We report results for the classifier training setting as well in Appendix Table~\ref{tab:no_hc_ablations_cls}. We find trends consistent with our main results (in Table~\ref{tab:cls_ft_results}), where we see a general trend of performance gains with human context over no author's context for \hlunroll, and only person-level tasks benefitting in case of non-\hulm models (i.e., \llama and \lhlc), with a common exception for stance and sentiment tasks across the table with no statistical significance observed.

\begin{table*}[h!]
    \centering
    \setlength{\tabcolsep}{5pt} 
    \renewcommand{\arraystretch}{1.1} 
    \begin{small}
    \begin{adjustbox}{width=\textwidth} 
    \begin{tabular}{l|l|cccccc|c|c}

        \hline
        
        \tbf{} & \tbf{} 
        & \multicolumn{6}{c|}{\tbf{Document-Level Labels}} 
        & \multicolumn{2}{c}{\tbf{Person-Level Labels}} \\


        \tbf{} & \tbf{} 
        & \multicolumn{6}{c|}{\tbf{Cls ($F1$)}} 
        & \multicolumn{1}{c|}{\tbf{Cls ($F1$)}} 
        & \multicolumn{1}{c}{\tbf{Reg ($r$)}}\\
        
        \tbf{Model} & \textbf{FT style} 
        & \tbf{Movie} & \tbf{Business} 
        & \tbf{Book} & \tbf{Elec}
        & \tbf{Stn} & \tbf{Sent} 
        & \tbf{Occ} 
        & \tbf{Age} \\
        
        \hline

        \llama  & TFT 
        & 66.40 & 70.93 & 65.32 & 65.26
        & 71.37 & 76.44
        & 52.45
        & 0.882 \\

        \lhlc  & TFT 
        & 66.10 & 70.93 & 67.41 & 66.44$^o$
        & \textbf{73.65}$^{}$ & \textbf{76.60} 
        & 51.69
        & 0.881$^o$ \\


        \hline

        \llama{} & \bf \huft 
        & 67.52 & \textbf{74.36}$^{}$ & 69.22 & \textbf{68.84}$^{}$
        & 69.09 & 76.07 
        & 52.28
        & \textbf{0.916}$^{}$  \\

        \lhlc{} & \bf \huft 
        & \textbf{67.70}$^{}$ & 73.56 $\hat{}$ & 69.12 & 68.16 $\hat{}$
        & 72.41 $\hat{}$ & 75.77 
        & 54.17
        & 0.915 \\
        
        \hlunroll & \bf \huft 
        & 67.05$^{*\dagger}$ & 73.36$^*$ & \textbf{69.43$^*$} & 68.22$^*$
        & 70.77$^*$ & 75.25 
        & \textbf{57.50$^{*\dagger}$}
        & \textbf{0.916$^*$} \\

        \hline
    \end{tabular}
    \end{adjustbox}
    \end{small}
    \caption{ Evaluating the effect of our LHLC data corpus on QLoRA based HuFT. Results are reported in weighted F1 for classification (cls) tasks and in pearson r for regression (reg) task. Bold indicates best in column. We indicate statistical significant ($p < .05$) difference in results between: \lhlc HuFT and \hlunroll HuFT with $^\dagger$, \lhlc TFT and \hlunroll HuFT using $^*$, \lhlc TFT and \llama TFT with $^0$, and \lhlc HuFT and \llama HUFT using $\hat{}$.
    }
    \label{tab:lhlc_qlora_results}
\end{table*}

\begin{table*}[h!]
    \centering
    \setlength{\tabcolsep}{5pt} 
    \renewcommand{\arraystretch}{1.1} 
    \begin{small}
    \begin{adjustbox}{width=\textwidth} 
    \begin{tabular}{l|l|cccccc|c|c}

        \hline
        
        \tbf{} & \tbf{} 
        & \multicolumn{6}{c|}{\tbf{Document-Level Labels}} 
        & \multicolumn{2}{c}{\tbf{Person-Level Labels}} \\


        \tbf{} & \tbf{} 
        & \multicolumn{6}{c|}{\tbf{Cls ($F1$)}} 
        & \multicolumn{1}{c|}{\tbf{Cls ($F1$)}} 
        & \multicolumn{1}{c}{\tbf{Reg ($r$)}}\\
        
        \tbf{Model} & \textbf{FT style} 
        & \tbf{Movie} & \tbf{Business} 
        & \tbf{Book} & \tbf{Elec}
        & \tbf{Stn} & \tbf{Sent} 
        & \tbf{Occ} 
        & \tbf{Age} \\
        
        \hline

         \llama & TFT
    & 57.99$^*$ & 65.99$^*$ & 59.90$^*$ & 59.06$^*$
    & 64.29 & 62.01
    & 47.48
    & 0.826 \\

      \llama & HuFT
    & 55.06 & 63.37 & 57.80 & 57.61
    & 62.96 & 61.40 
    & 53.59$^*$
    & 0.853$^*$ \\

    \hline

    \lhlc & TFT
    & 58.31$^*$ & 65.64$^*$ & 61.18$^*$ & 60.22$^*$
    & \textbf{69.25} & \textbf{63.17} 
    & 50.61
    & 0.835 \\

     \lhlc & HuFT
    & 55.43 & 63.57 & 59.36 & 57.54
    & 66.37 & 62.38 
    & 54.01$^*$
    & 0.855$^*$ \\

    \hline

    \hlunroll & TFT
    & 58.99 & 66.15 & 61.15 & 60.15
    & 68.21 & 62.72 
    & 47.38
    & 0.839 \\

      \hlunroll & \textbf{HuFT}
    & \textbf{58.48} & \textbf{66.80$^*$} & \textbf{61.88$^*$} & \textbf{60.93$^*$}
    & 67.08 & 62.93 
    & \textbf{54.73$^*$}
    & \textbf{0.858$^*$} \\

        \hline
    \end{tabular}
    \end{adjustbox}
    \end{small}
    \caption{Evaluating the effect of adding author's historical language in the setting for classifier training with human context by comparing with TFT setting (i.e., no human context). Results are reported in weighted F1 for classification (cls) tasks and in pearson r for regression (reg) task. Bold indicates best in column. $^*$ indicates statistical significance $p < .05$ for each model separately between their respective TFT and HuFT results, with $^*$ marked on the better result between the two.
    }
    \label{tab:no_hc_ablations_cls}
\end{table*}

\begin{table*}[h!]
    \centering
    \setlength{\tabcolsep}{5pt} 
    \renewcommand{\arraystretch}{1.1} 
    \begin{small}
    \begin{adjustbox}{width=\textwidth} 
    \begin{tabular}{l|l|cccccc|c|c}

        \hline
        
        \tbf{} & \tbf{} 
        & \multicolumn{6}{c|}{\tbf{Document-Level Labels}} 
        & \multicolumn{2}{c}{\tbf{Person-Level Labels}} \\


        \tbf{} & \tbf{} 
        & \multicolumn{6}{c|}{\tbf{Cls ($F1$)}} 
        & \multicolumn{1}{c|}{\tbf{Cls ($F1$)}} 
        & \multicolumn{1}{c}{\tbf{Reg ($r$)}}\\
        
        \tbf{Model} & \textbf{FT style} 
        & \tbf{Movie} & \tbf{Business} 
        & \tbf{Book} & \tbf{Elec}
        & \tbf{Stn} & \tbf{Sent} 
        & \tbf{Occ} 
        & \tbf{Age} \\
        
        \hline

        \llama  & TFT 
        & 66.40 & 70.93 & 65.32 & 65.26
        & 71.37 & 76.44
        & 52.45
        & 0.882 \\

         \llama{} & \bf \huft 
        & \doublespace{}67.52$^*$ & \textbf{\doublespace{}74.36$^*$} & \doublespace{}69.22$^*$ & \textbf{\doublespace{}68.84$^*$}
        & 69.09 & 76.07 
        & 52.28
        & \textbf{\doublespace{}0.916$^*$}  \\

        \hline

        \lhlc  & TFT 
        & 66.10 & 70.93 & 67.41 & 66.44
        & \textbf{\doublespace{}73.65$^*$} & 76.60 
        & 51.69
        & 0.881 \\

          \lhlc{} & \bf \huft 
        & \textbf{\doublespace{}67.70$^*$} & \doublespace{}73.56$^*$ & \doublespace{}69.12$^*$ & \doublespace{}68.16$^*$
        & 72.41 & 75.77 
        & 54.17
        & \doublespace{}0.915$^*$ \\

        \hline

        \hlunroll & TFT 
        & 66.19 & 70.82 & 67.24 & 65.85
        & 72.32 & \textbf{\doublespace{}78.17$^*$}
        & 52.02
        & 0.884 \\

        \hlunroll & \bf \huft 
        & \doublespace{}67.05$^*$ & \doublespace{}73.36$^*$  & \textbf{\doublespace{}69.43$^*$} & \doublespace{}68.22$^*$
        & 70.77 & 75.25 
        & \textbf{\doublespace{}57.50$^*$}
        & \textbf{\doublespace{}0.916$^*$} \\

        \hline
    \end{tabular}
    \end{adjustbox}
    \end{small}
    \caption{Evaluating the effect of adding author's historical language in the QLoRA setting with human context (HuFT) by comparing with QLoRA TFT setting (i.e., no human context). Results are reported in weighted F1 for classification (cls) tasks and in pearson r for regression (reg) task. Bold indicates best in column. $^*$ indicates statistical significance $p < .05$ for each model separately between their respective TFT and HuFT results, with $^*$ marked on the better result between the two.
    }
    \label{tab:no_hc_ablations_qlora}
\end{table*}

\begin{table*}[h!]
    \centering
    \begin{tabular}{l|l|cc}

        \hline
        
        \tbf{} & \tbf{} 
        & \multicolumn{2}{c}{\tbf{Document-Level Labels}} \\


        \tbf{} & \tbf{} 
        & \multicolumn{2}{c}{\tbf{Cls ($F1$)}} \\
        
        \tbf{Model} & \textbf{\makecell{Human Context (HC)}}
        & \tbf{Business} 
        & \tbf{Elec}
        \\
        
        \hline

         \llama-Instruct & No HC
    & 59.17 & 59.72 
   
     \\

    \hline

    \llama-Instruct & Full HC
    & 60.34 & \textbf{61.16} 
  
     \\

    \llama-Instruct & Limited HC (500 words)
    & 60.29 & \textbf{61.17}
   
    \\

    \llama-Instruct & Limited HC (250 words)
    & \textbf{\doublespace{}60.95$^*$}  & 61.10 \\

    \llama-Instruct & Labeled HC 
    & \doublespace{}60.63$^{o}$  & \doublespace{}56.44$^{o\dagger}$
    
    \\

        \hline
    \end{tabular}
    \caption{Results with ablations for directly including human context (HC) in prompting using \llama-Instruct 3.1 8B model over selected downstream tasks. Results are reported in weighted F1 for classification (cls) tasks. Bold indicates best in column. We indicates statistical significance ($p < .05$) difference in results between: best in column and Full HC using $^*$, Labeled HC and No HC with $^o$, and Labeled HC and best in column using $^\dagger$.
    }
    \label{tab:zero_shot_prompt_ablations}
\end{table*}

\subsection{Prompting}
\label{app:prompt}

We use  llama3.1-8b-Instruct-hf model and the vLLM framework~\cite{kwon2023efficient} for our prompting experiments. The experiments were run on an RTX A6000
48GB GPU.
For prompting, we use 2 different methods (with and without author context). For Business and Electronics, we did additional experiments (limited author context, labeled author context). In limited author context, we limit the author context and take first [250, 500] words. Prompt details are given in Table~\ref{tab:prompt-templates}.

\begin{table*}[t]
\centering
\small
\begin{tabular}{@{}p{2.5cm}p{14.5cm}@{}}
\toprule
\textbf{Task} & \textbf{Prompt Template} \\
\midrule
Stance Topic &
\begin{tabular}[t]{@{}l@{}}
Identify the stance of the given target text
towards \{topic\}. Select one of the three: In Favor, or Against, or Neutral. \\
Here is the target text: \\
\{text\} \\
Do not include any extra information.
\end{tabular} \\
\midrule
Stance Topic with Author Context &
\begin{tabular}[t]{@{}l@{}}
Here is a list of the previous messages written by the person in chronological order to learn more about the person: \\
\{messages\} \\
Identify the stance of the given target text
towards \{topic\}. Select one of the three: In Favor, or Against, or Neutral. \\
Here is the target text: \\
\{text\} \\
Do not include any extra information.
\end{tabular} \\
\midrule
Sentiment &
\begin{tabular}[t]{@{}l@{}}
Identify the sentiment of the given target text. Select one of the three: Positive, or Negative, or Neutral. \\
Here is the target text: \\
\{text\} \\
Do not include any extra information.
\end{tabular} \\
\midrule
Sentiment with Author Context &
\begin{tabular}[t]{@{}l@{}}
Here is a list of the previous messages written by the person in chronological order to learn more about the person: \\
\{messages\} \\
Identify the sentiment of the given target text. Select one of the three: Positive, or Negative, or Neutral. \\
Here is the target text: \\
\{text\} \\
Do not include any extra information.
\end{tabular} \\
\midrule
Reviews Classification &
\begin{tabular}[t]{@{}l@{}}
Determine the star rating (from 1 to 5) that best reflects the following \{review\} and answer strictly as only one\\
from [1,2,3,4,5].
\\
\{review\}:\\
\{text\}
\\
Do not include any extra information.
\end{tabular} \\
\midrule
Reviews Classification with Author Context &
\begin{tabular}[t]{@{}l@{}}
Here is a list of the previous \{category\}  written by the person in chronological order to learn more about the person:
\\
\{messages\} \\
Determine the star rating (from 1 to 5) that best reflects the following \{review\} and answer strictly as only one\\
from [1,2,3,4,5].  \\
\{review\}:\\
\{text\}\\
Do not include any extra information.
\\

\end{tabular} \\
\midrule
Reviews Classification with labeled Author Context &
\begin{tabular}[t]{@{}l@{}}
Here is a list of the previous \{category\}  written by the person in chronological order and the corresponding rating \\ that they gave:
\\
\{review rating pairs\} \\
Determine the star rating (from 1 to 5) that best reflects the following \{review\} and answer strictly as only one\\
from [1,2,3,4,5].  \\
\{review\}:\\
\{text\}\\
Do not include any extra information.
\\
\end{tabular} \\
\midrule
Job Classification &
\begin{tabular}[t]{@{}l@{}}
Given a list of messages written by a person, predict their most relevant job category as only one of the following: \\
Education, Student, Technology, Arts, Communications-Media. \\
Here is a list of the person’s written messages in chronological order: \\
\{messages\} \\
Now predict the person's job category. \\
Do not include any extra information.
\end{tabular} \\
\midrule
Age Estimation &
\begin{tabular}[t]{@{}l@{}}
Given a list of messages written by a person, estimate the person's age. \\
Here is a list of the person’s written messages in chronological order: \\
\{messages\} \\
Now just give the person's age as a real valued number without any explanation. \\
Give only the age value between 0 to 100 and no other text.
\end{tabular} \\
\bottomrule
\end{tabular}
\caption{topic = [Hillary Clinton, atheism, feminism, legalization of abortion, climate change as a real concern], review = [Movie review for Movies, Business review for Business, review for books and electronics], category = [movie reviews, reviews for different businesses, reviews for books available on amazon, reviews for electronic products available on amazon]. \{review rating pairs\}, \{messages\} are separated by a line. For "limited author context" we limit the \{messages\} to [250, 500] words.}
\label{tab:prompt-templates}
\end{table*}

\vspace{1em}
\noindent





\subsection{Qualitative Analysis}
\label{app:qa_process}

We qualitatively evaluate the effect of including human context beyond the experimental analysis discussed by manually looking at downstream task predictions using a structured qualitative analysis process. We compare the predictions from the best model in the \huft setting versus the best model in the \tft setting in two scenarios: a) \huft predicts correctly and \tft predicts incorrectly, and b) vice-versa. Next, we sort the instances by the amount of author's historical context used in these cases and manually review to find where this human context proved to be helpful (i.e., first scenario), and where the human context may turn out to be ambiguous and thus hurting the model's judgement. 

\begin{table*}
    \centering
    \setlength{\tabcolsep}{5pt}
    \footnotesize
    \begin{tabular}{p{1.0cm}|p{3.0cm}|p{1.2cm}|p{1.2cm}|p{1.2cm}|p{3.3cm}|p{3.2cm}}
    \hline
    \textbf{Task} &
    \textbf{Target Text} &
    \textbf{Gold} &
    \textbf{With HC} &
    \textbf{No HC} &
    \textbf{Selected Historical Language} &
    \textbf{Why History Helps/Hurts} \\
    \hline

    Stance Pred. (Abortion) &
    \texttt{@\_msthirdward} \texttt{@BuhayIpaglaban} \texttt{@blueskies366}
    It's never been about ``life'' has it?
    \negative{``\#Prolife'' are OK with women dying}. &
    \positive{AGAINST} &
    \positive{AGAINST} &
    \negative{FOR} &
    [\dots] It's never been about ``life'' has it?
    \positive{``\#Prolife'' are OK with women dying}. [\dots]
    The culture of death is \positive{\#prolife who deny pregnant people healthcare}. [\dots]
    thank you again to show that you are \positive{AGAINST CONTRACEPTION};
    ``\#Prolife'' = a code word for \positive{misogyny}. [\dots]
    Tell me more about how \positive{\#antichoice laws kill 47,000 women A YEAR}! [\dots]
    Remember \positive{Savita, killed by Irish ``\#prolife'' laws}?
    \positive{Repeal the 8th}. [\dots] &
    Without historical context, the model interprets the tweet as \negative{ambiguous or rhetorical}.
    User history contains strong, repeated \positive{anti-prolife} and \positive{pro-choice} language, which \positive{disambiguates stance}. \\
    \hline

    Stance Pred. (Atheism) &
    I \negative{\#DenounceHarper} for refusing to include
    \negative{family planning} in foreign aid even though spending \$1 could save \$6
    \negative{\#wherestheFP} &
    \positive{FOR} &
    \positive{FOR} &
    \negative{AGAINST} &
    Enough sneaking around, bringing in \positive{religious groups into public schools} to access our kids in secrecy. [\dots]
    I wonder if \positive{anti-choice advocates denying accurate SexED} ever consider the harm they cause. [\dots]
    How many can you fit in your womb --- \positive{mocking reproductive control arguments}. [\dots]
    \positive{Anti-choice actors never try to police men over sex}, but punishing women puts them into a frenzy. &
    The target criticizes a political decision but does not explicitly mention \negative{religion}.
    History shows persistent opposition to \positive{religious influence in policy/education}, enabling the correct inference. \\
    \hline

    Stance Pred. (Abortion) &
    I \negative{\#DenounceHarper} for refusing to include
    \negative{family planning} in foreign aid even though spending \$1 could save \$6
    \negative{\#wherestheFP} &
    \positive{FOR} &
    \positive{FOR} &
    \negative{AGAINST} &
    \positive{family planning as part of foreign aid}. [\dots]
    Sneaking \positive{religious groups into public schools} to access kids in secrecy. [\dots]
    Pregnant teens denied accurate \positive{SexED} suffer long-term consequences. [\dots]
    Talking about fertility clinics and rescue rhetoric highlights \positive{anti-choice framing around women's bodies and wombs}. [\dots] &
    Without context, the model associates foreign-aid criticism with \negative{conservative framing}.
    History consistently links family planning/contraception with \positive{women’s health} and \positive{anti-restriction advocacy}. \\
    \hline

    \end{tabular}
    \caption{Stance prediction qualitative examples where incorporating human context (HC) changes the model’s interpretation and improves stance inference.}
    \label{tab:stance_examples_hc}
\end{table*}

\begin{table*}
    \centering
    \setlength{\tabcolsep}{5pt}
    \footnotesize
    \begin{tabular}{p{1.0cm}|p{3.0cm}|p{1.2cm}|p{1.2cm}|p{1.2cm}|p{3.3cm}|p{3.2cm}}
    \hline
    \textbf{Task} &
    \textbf{Target Text} &
    \textbf{Gold} &
    \textbf{With HC} &
    \textbf{No HC} &
    \textbf{Selected Historical Language} &
    \textbf{Why History Helps/Hurts} \\
    \hline

    Stance Pred. (Feminism) &
    So if a women earns money then it's HERS to KEEP?? \negative{\#feminismiscruelty} &
    \negative{AGAINST} &
    \positive{FOR} &
    \negative{AGAINST} &
    [\dots] So if a woman earns money then it's HERS to KEEP?? [\dots]
    UGH I am SICK to DEATH of this whole \negative{women wanting ``equality''} thing. [\dots]
    Drawing attention to women's health issues is framed as
    \negative{evidence that feminism is harmful or cruel}. [\dots]
    A woman wanting to be equal to a man is portrayed as
    \negative{something outrageous and unnatural}. [\dots]
    \negative{Feminism is characterized as a hate group rather than an equality movement}. [\dots] &
    History is repetitive and emotionally charged. The model over-indexes on
    \negative{sarcasm/hostile tone}, which can \negative{flip polarity}
    despite a consistent anti-feminist intent. \\
    \hline

    Stance Pred. (Abortion) &
    A prochoice advocate but circumcise ur baby? Fucking hypocrite!
    \negative{\#circumcision} \negative{\#humanrights} &
    \negative{AGAINST} &
    \positive{FOR} &
    \negative{AGAINST} &
    [\dots] A \negative{pro-choice advocate framed as morally hypocritical} for supporting circumcision. [\dots]
    If a man wants abortion but the woman wants to keep it,
    \negative{he should not be responsible for child support}. [\dots]
    \negative{Modern feminism is described as irrational and self-serving}. [\dots]
    Violence against women is contrasted with
    \negative{claims that feminism focuses on trivial cultural issues}. [\dots]
    Broader grievances about gender roles dominate the historical context. [\dots] &
    History spans multiple issues (abortion, feminism, gender roles), causing
    \negative{topic drift}. The model generalizes from broader ideology instead of
    \positive{anchoring to the abortion-specific claim}. \\
    \hline

    \end{tabular}
    \caption{Stance prediction qualitative examples where incorporating human context (HC) does not help: history can amplify tone or introduce topic drift, leading to incorrect predictions.}
    \label{tab:stance_examples_hc_nohelp}
\end{table*}

\begin{table*}
    \centering
    \setlength{\tabcolsep}{5pt}
    \footnotesize
    \begin{tabular}{p{1.0cm}|p{3.0cm}|p{1.2cm}|p{1.2cm}|p{1.2cm}|p{3.3cm}|p{3.2cm}}
    \hline
    \textbf{Task} &
    \textbf{Target Text} &
    \textbf{Gold} &
    \textbf{With HC} &
    \textbf{No HC} &
    \textbf{Selected Historical Language} &
    \textbf{Why History Helps/Hurts} \\
    \hline

    Review Pred. (Amazon) &
    If you have a hard time falling asleep, \negative{buy this DVD --- you won't regret it}.
    You'll be asleep before it ends.
    The only way to see how it ends is to fast forward to the last 10 minutes and you will still not know what the plot is about.
    Even if you manage to keep your eyes open throughout the entire DVD. &
    \negative{0.0} &
    \negative{0.0} &
    \positive{4.0} &
    [\dots] I did not care for this movie \dots\ reading cost nothing but your time and eyesight \dots\ the DVD does not disclose anything more
    than you can find out for yourself! (\negative{0.0}) [\dots]
    the movie is \positive{extremely boring}, every scene is predictable \dots\ \positive{Don't waste your time}
    like I did. (\negative{0.0}) [\dots]
    this is a \positive{very poorly written} movie \dots\ the preview makes you believe this might be funny ---
    it is \positive{far from being funny}. (\negative{0.0}) [\dots]
    The most entertaining movie I have seen in a very long time \dots\ It is a must see movie.
    (\positive{4.0}) [\dots]
    The script was well written, the acting superb \dots\ one of the funniest movies I've seen in a long time. (\positive{4.0}) [\dots] &
    Without historical context, the sarcastic \negative{``you won't regret it''} reads as genuine praise.
    The user's history shows a consistent pattern where sleep/boredom metaphors co-occur with
    \positive{``extremely boring''} and \positive{``Don't waste your time''}, clarifying this as a negative review. \\
    \hline

    Review Pred. (Amazon) &
    What a \negative{horrifying movie}! I wish Tim would have gotten an Oscar for this movie. &
    \positive{4.0} &
    \positive{4.0} &
    \negative{0.0} &
    [\dots] What an all-star cast that all \positive{performed excellent}!
    The ending was great as well.
    \positive{Sean Penn did an incredible job} on this film. (\positive{3.0}) [\dots]
    The script was \positive{well written}, the acting \positive{superb};
    this movie is one of the funniest I've seen in a long time. (\positive{4.0}) [\dots]
    This and Revenge have been Kevin Costner's \positive{best performances}.
    Amazing script, directing, and acting. (\positive{4.0}) [\dots]
    By contrast, clearly negative reviews contain unambiguous dismissal such as
    \positive{``extremely boring''}, \positive{``don't waste your time''},
    and \positive{``very poorly written''}, all paired with low ratings (\negative{0--1}). [\dots] &
    Without historical context, \negative{``horrifying''} is treated as a negative cue.
    The user's history makes clear that strong sentiment words are not decisive, while consistent emphasis on
    \positive{performance/acting praise} aligns with high ratings. \\
    \hline

    \end{tabular}
    \caption{Amazon reviews qualitative examples (with vs.\ without human context). Positive examples where human context helps.}
    \label{tab:amazon_examples_pos}
\end{table*}

\begin{table*}
    \centering
    \setlength{\tabcolsep}{5pt}
    \footnotesize
    \begin{tabular}{p{1.0cm}|p{3.0cm}|p{1.2cm}|p{1.2cm}|p{1.2cm}|p{3.3cm}|p{3.2cm}}
    \hline
    \textbf{Task} &
    \textbf{Target Text} &
    \textbf{Gold} &
    \textbf{With HC} &
    \textbf{No HC} &
    \textbf{Selected Historical Language} &
    \textbf{Why History Helps/Hurts} \\
    \hline

    Review Pred. (Amazon) &
    \positive{I have DVD's from other region zones and PAL DVD's and they all play on my Blu Ray\ldots\ why wouldn't this one?}
    Someone who has gotten this and is playing it on a US Blu Ray player, please respond! &
    \positive{4.0} &
    \negative{0.0} &
    \positive{4.0} &
    [\dots] I had the soundtrack on Colgems records and \negative{loved the music and dialog};
    when I saw the movie was on DVD, I grabbed it!
    \negative{Great nostalgia}. (\positive{3.0}) [\dots]
    I read people complain that this is long and boring, but this documentary is \negative{like no other in rock history}.
    Morrison was a \negative{genius}, complicated and tortured, and this film \negative{opened my eyes}.
    (\positive{4.0}) [\dots] &
    The review is a neutral, practical question about disc compatibility, not a complaint.
    Without history, the model \positive{treats it as informational} and predicts a positive rating.
    With history, the model \negative{expects explicit enthusiasm} and overcorrects, interpreting the restrained tone as dissatisfaction. \\
    \hline

    Review Pred. (Amazon) &
    \positive{I purchased this new from media\_distributors but I made the mistake of not looking at it right away.}
    It looked like it was new and the video was labeled Blonde Crazy\ldots\ When I played it, it turned out to be Rumpole.
    I wrote an email to Media Distributors after posting this, and they replaced my video.
    \positive{Outstanding}. &
    \negative{0.0} &
    \positive{4.0} &
    \negative{0.0} &
    [\dots] A collection of way overpriced videos that you can get in bargain bins for \$5 each. (\negative{1.0}) [\dots]
    The Alpha version is of \negative{very poor video quality}, though the underlying film may have been interesting. (\negative{1.0}) [\dots]
    This is not a great film, but it is a \negative{fun film} and interesting to see Robert Mitchum in his first feature. (\positive{2.0}) [\dots]
    The Alpha version of Millie is \negative{exceptional} with \negative{outstanding print quality}. (\positive{3.0}) [\dots] &
    The review reports a shipping error that was resolved, but the overall experience is still framed around a mistake.
    Without history, the model \positive{anchors on the problem-focused content} and predicts a low rating.
    With history, prior instances of praising releases/distributors after logistics make the model \negative{overweight a single positive cue} (``Outstanding'') and inflate the rating. \\
    \hline

    \end{tabular}
    \caption{Amazon reviews qualitative examples (with vs.\ without human context). Negative examples where human context does not help.}
    \label{tab:amazon_examples_neg}
\end{table*}


\begin{table*}
    \centering
    \setlength{\tabcolsep}{5pt}
    \footnotesize
    \begin{tabular}{p{1.0cm}|p{3.0cm}|p{1.2cm}|p{1.2cm}|p{1.2cm}|p{3.3cm}|p{3.2cm}}
    \hline
    \textbf{Task} &
    \textbf{Target Text} &
    \textbf{Gold} &
    \textbf{With HC} &
    \textbf{No HC} &
    \textbf{Selected Historical Language} &
    \textbf{Why History Helps/Hurts} \\
    \hline

    Review Pred. (Yelp) &
    Do you have a heart condition, high cholesterol, diabetes, over-weight, or anything like that?
    Well, here is your swan song that's worth the further \negative{damage} --- the Daily Dog at Lady Di's.
    A deep fried hot dog stuffed with gooey cheese, wrapped in crispy bacon.
    Eat one, you'll find nirvana. Eat two, you'll find the \negative{ER}. &
    \positive{4.0} &
    \positive{4.0} &
    \negative{0.0} &
    [\dots] Oh dear laaawd! That cheeseburger on a pretzel bun with onion straws ---
    \positive{are you kidding me?!?} (\positive{3.0}) [\dots]
    Fast food is standard everywhere, but this Taco Bell always gets the order wrong.
    Mexican pizza with tomatoes and no cheese instead. (\negative{1.0}) [\dots]
    Chuck E Cheese has mediocre pizza, dirty salad bars, and chaos everywhere.
    This place has it all --- and not in a good way. (\negative{0.0}) [\dots] &
    In isolation, the exaggerated health warnings (\negative{``ER''}, \negative{``damage''}) read as criticism.
    The user’s history shows that this kind of sarcastic, over-the-top storytelling is how they express strong approval.
    With context, the humor and sensory detail signal a positive experience rather than a complaint. \\
    \hline

    Review Pred. (Yelp) &
    Hey Megabus, guess where I am right now? Yeah, that’s right --- I’m on Bolt Bus writing a review
    because \negative{Bolt Bus actually gave me the wifi they promised me}.
    Megabus, you are always Megalate, Megarude, and I will not be taking you anymore. &
    \negative{0.0} &
    \negative{0.0} &
    \positive{4.0} &
    [\dots] Their \positive{service is terrible}!
    I will \positive{NEVER recommend} them to anyone. (\negative{0.0}) [\dots]
    I’m giving this place 3 stars because of their \positive{horrible service}.
    The bagels are great, but the service leaves a bad taste. (\negative{2.0}) [\dots]
    I really wanted to enjoy the food and service, but they \positive{fell short} in both. (\negative{1.0}) [\dots]
    This truly is the best pizza ever! Everything tastes fresh and the staff are inviting. (\positive{4.0}) [\dots] &
    Without historical context, the review reads as enthusiastic because it highlights concrete benefits (\negative{wifi}) and praises a competitor, leading to a positive prediction.
    In the user’s history, genuine positives are direct and unambiguous, while sarcastic, insult-heavy rants like this one consistently map to 0-star ratings, disambiguating the intent as negative. \\
    \hline

    \end{tabular}
    \caption{Yelp reviews qualitative examples (with vs.\ without human context). Positive examples where human context helps. For these cases, \negative{red highlights in the target} indicate misleading cues without context, while \positive{green highlights in history} indicate disambiguating signals.}
    \label{tab:yelp_examples_pos}
\end{table*}

\begin{table*}
    \centering
    \setlength{\tabcolsep}{5pt}
    \footnotesize
    \begin{tabular}{p{1.0cm}|p{3.0cm}|p{1.2cm}|p{1.2cm}|p{1.2cm}|p{3.3cm}|p{3.2cm}}
    \hline
    \textbf{Task} &
    \textbf{Target Text} &
    \textbf{Gold} &
    \textbf{With HC} &
    \textbf{No HC} &
    \textbf{Selected Historical Language} &
    \textbf{Why History Helps/Hurts} \\
    \hline

    Review Pred. (Yelp) &
    I've never had Chinese food before, and thought to myself after the first bite,
    \positive{``wow this tastes like it was just cooked for me.''}
    \positive{This is the experience we had not once but every time we have ordered from Lingnan.} &
    \positive{4.0} &
    \negative{1.0} &
    \positive{4.0} &
    [\dots] The first thing that made me decide I would \negative{never eat here}
    was the smell and the condition of their kitchen.
    I was \negative{gagging}. (\negative{0.0}) [\dots]
    We were appalled to find they didn’t offer refried beans.
    We \negative{never ended up eating anything}. (\negative{1.0}) [\dots]
    This place must be stuck in the \negative{early 90's}.
    No debit, overpriced ATM --- \negative{no thanks}. (\negative{1.0}) [\dots]
    Despite the flies and pricing, the pulled pork sandwich was great,
    but \negative{a lot really pissed me off}. (\negative{2.0}) [\dots] &
    The review is clearly positive, with \positive{direct praise} and \positive{repeated satisfaction}.
    Without history, the model follows this surface sentiment.
    With history, frequent \negative{complaint-heavy reviews} set a low baseline, causing the model to
    \negative{discount brief praise} and predict too low. \\
    \hline

    Review Pred. (Yelp) &
    \positive{I love this place for some late night eats.}
    \positive{I would drive past Viva Burrito and go to El Potosino because they have bigger and better burritos.}
    Viva is dirty and the employees are lazy, but
    \positive{El Potosino has great California burritos with guacamole and is one of the best late night eats}
    for the \positive{price, quantity, and quality}. &
    \positive{3.0} &
    \negative{1.0} &
    \positive{3.0} &
    [\dots] I don’t understand why people still come to this restaurant ---
    the food has \negative{gone downhill} and is \negative{overpriced}. (\negative{1.0}) [\dots]
    Fast and ok, but the new items were \negative{soggy} and I \negative{wouldn’t recommend} them.
    (\negative{2.0}) [\dots]
    This place is \negative{terrible}. Bad service, bad food --- I \negative{wouldn’t recommend it at all}.
    (\negative{0.0}) [\dots]
    I do love this place, \negative{but the service is terrible} and there’s \negative{no urgency}.
    (\negative{0.0}) [\dots] &
    The user criticizes one restaurant but is clearly \positive{endorsing another}.
    Without history, the model follows that intent.
    With history, frequent past \negative{complaints} make the model overreact to negative words and
    \negative{overlook the recommendation}. \\
    \hline

    \end{tabular}
    \caption{Yelp reviews qualitative examples (with vs.\ without human context). Negative examples where human context does not help. For these cases, \negative{red highlights in history} indicate misleading priors, while \positive{green highlights in the target} indicate the intended sentiment.}
    \label{tab:yelp_examples_neg}
\end{table*}

\begin{table*}
    \centering
    \setlength{\tabcolsep}{5pt}
    \footnotesize
    \begin{tabular}{p{1.0cm}|p{3.0cm}|p{1.2cm}|p{1.2cm}|p{1.2cm}|p{3.3cm}|p{3.2cm}}
    \hline
    \textbf{Task} &
    \textbf{Target Text} &
    \textbf{Gold} &
    \textbf{With HC} &
    \textbf{No HC} &
    \textbf{Selected Historical Language} &
    \textbf{Why History Helps/Hurts} \\
    \hline

    Occ &
    {“People talk about all this unused potential in the world, but the greatest untapped power is the human mind…” [TFT → Arts] [...]
    \negative{“Language is fascinating — the brain doesn’t process words the way we think it does.” [TFT → Arts]} [...]
    \negative{“If we limit what ideas people are exposed to, we limit the future itself.” [TFT → Arts]} [...]
    "... homeward bound :: need my hermit shell back ::" [TFT → Technology]
    } &
    \positive{Tech} &
    \positive{Tech} &
    \negative{Arts} ["Arts", 35, "Tech.", 24, "Student", 13, "Comm.-Media", 6, "Edu", 4] &
    i am mack, i will some day become great [\dots]
    People talk about unused \positive{CPU power} [\dots]
    the human brain [\dots]
    Discussion of banning \positive{evolution from science classes} [\dots]
    Reactions to \positive{NASA discoveries} and scientific progress [\dots]
    homeward bound :: need my hermit shell back [\dots] &
    Without history, expressive and philosophical phrasing is easy to read as \negative{artistic or abstract writing}.
    Across the user’s history, repeated references to \positive{computing metaphors},
    \positive{science education}, and \positive{technology--policy reasoning}
    clarify a technology-oriented mindset rather than an arts identity. \\
    \hline

    Occ &
    
    {
    “Played \negative{Doom III} with Bob yesterday from 2:00 until dark. At one point I noticed strange twinkling lights coming from the bottom of his blinds. Turns out it was the sun. We had no idea what time it was.” [TFT → Tech.] [...]
    “Now I'm off to \negative{Future Shop}. The \negative{Zoom button} on my \negative{camera} has become stuck, and I need it repaired. Wish me luck.” [TFT → Tech.] [...]
    “The last few days of school annoy me. Monday and Tuesday are the last official days, but attendance is extremely sporadic — people would much rather study for exams or play in the sun than sit in classes. Besides, the teachers have all but given up teaching; classes consist of movies, maybe a pretest or two.” [TFT → Student] [...]
    }
    &
    \positive{Student} &
    \positive{Student} &
    \negative{Tech} ["Tech.", 25, "Student", 17, "Edu.", 17, "Arts", 14, "Comm.-Media", 10]  &
    Last official day of \positive{school} [\dots] had a \positive{chemistry final} [\dots]
    Success in my \positive{math final} and \positive{biology provincial} [\dots]
    \negative{Currently very pleased with Future Shop} [\dots]
    \positive{Classes} are completely done [\dots]
    \positive{Exams} start tomorrow [\dots]
    \negative{play some Super Nintendo} [\dots] &
    Without broader history, isolated mentions of tools or hobbies push the prediction toward \negative{Technology}.
    The full history shows a consistent focus on \positive{classes}, \positive{exams}, and \positive{academic routines},
    establishing a clear student-centered identity. \\
    \hline

    \end{tabular}
    \caption{User-level job prediction examples (with vs.\ without human context). Target text is omitted for user-level classification; history provides the primary signal.}
    \label{tab:job_examples_pos}
\end{table*}

\begin{table*}[t]
\centering
\setlength{\tabcolsep}{5pt}
\footnotesize
\begin{tabular}{p{1.0cm}|p{3.0cm}|p{1.2cm}|p{1.2cm}|p{1.2cm}|p{3.3cm}|p{3.2cm}}
\hline
\textbf{Task} &
\textbf{Target Text} &
\textbf{Gold} &
\textbf{With HC} &
\textbf{No HC} &
\textbf{Selected Historical Language} &
\textbf{Why History Helps/Hurts} \\
\hline

Occ. &
 &
\positive{Tech.} &
\negative{Student} &
\positive{Tech.} &
study F.6 … subjects choices are: Physics, Biology / \negative{Pure Mathematics}, \negative{Computer Studies} … A-level syllabus is very demanding [...] 
release of \negative{HKCEE result} … procedure for applying F.6 [...] 
had a \negative{chemistry final} … success in my \negative{math final} [...] 
decided to write a polling web page using \positive{ASP.NET} … learned \positive{stored procedures} for \positive{SQL Server} [...] 
downloaded \positive{Visual C\# Express} … mixing .NET and native code [...] &
The history mixes \negative{strong school-centered signals} (exams, subject selection, results) with \positive{clear hands-on technical work}. When history is used, the model overweights repeated school references and collapses the user into a \negative{Student} identity. Without history, it relies more on concrete development activity (\positive{ASP.NET, C\#, SQL}), aligning better with \positive{Tech.}. \\
\hline

Occ. &
 &
\positive{Student} &
\negative{Tech.} &
\positive{Student} &
\negative{3rd party web browser for any modded Xbox} [...] HTML 4.0 support, HTTP 1.1, Javascript support [...] 
prefer \negative{Mozilla/Firefox within Xbox Linux} [...] stand-alone app launchable from any dashboard [...] 
\negative{Xbox modding tutorials} [...] release date of \negative{Microsoft Windows XP Service Pack 2} [...] 
played around with \negative{technical betas} [...] security dashboard, spyware, pop-ups [...] 
\positive{Ok so you wana play HALO xbox online} [...] step-by-step setup instructions [...] 
\positive{simple and easy to use}, free app [...] learning-oriented walkthrough [...] 
\positive{Really cool forum of a friend of mine} [...] perfect if you’re into halo [...] 
\positive{sharing screenshots} and experiments [...] curiosity-driven tinkering rather than deployment [...] &
The history contains many \negative{advanced technical terms and tools}, which makes the user appear professionally technology-oriented. However, these posts are mostly framed around \positive{learning, gaming, and experimentation}, with tutorial-style explanations and hobbyist motivation. Using history amplifies the technical surface cues and misleads the prediction away from the user’s true role as a \positive{Student}. \\
\hline

\end{tabular}
\caption{Negative examples where adding human context (HC) can mislead predictions. Target text is intentionally left blank.}
\label{tab:neg_examples_history_hurts}
\end{table*}

\end{document}